\documentclass[lettersize,journal]{IEEEtran}
\usepackage{amsmath,amsfonts}
\usepackage{algorithmic}
\usepackage{algorithm}
\usepackage{array}
\usepackage[caption=false,font=normalsize,labelfont=sf,textfont=sf]{subfig}
\usepackage{textcomp}
\usepackage{stfloats}
\usepackage{url}
\usepackage{verbatim}
\usepackage{graphicx}
\usepackage{hyperref}
\usepackage{cite}
\usepackage{tikz}
\usepackage{booktabs}
\usepackage{nth}
\newcommand{\loss}{\mathcal{L}}
\newcommand{\contentloss}{\loss_{c}}
\newcommand{\styleloss}{\loss_{s}}
\newcommand{\layernumber}{N_l}
\newcommand{\inputcontent}{I_c}

\begin{document}

\title{Inversion-Free Style Transfer with Dual Rectified Flows}

\author{Yingying Deng,
Xiangyu He,
Fan Tang,~\IEEEmembership{Member,~IEEE,}
Weiming Dong,~\IEEEmembership{Member,~IEEE,}
Xucheng Yin,~\IEEEmembership{Senior Member,~IEEE}
        % <-this % stops a space
% \thanks{This work was supported in part by the Beijing Natural Science Foundation under No.Z231100005923033. (Corresponding author: Xucheng Yin)}
\thanks{Yingying Deng and Xucheng Yin (Corresponding author) are with the Department of Computer Science and Technology,
University of Science and Technology Beijing, Beijing 100083, China
(e-mail: yingying.deng@ustb.edu.cn; xuchengyin@ustb.edu.cn)}
\thanks{Xiangyu He and Weiming Dong are with the Institute of Automation, Chinese Academy of Sciences, Beijing 100190, China (e-mail: hexiangyu17@mails.ucas.edu.cn; weiming.dong@ia.ac.cn)}
\thanks{Fan Tang is with the School of Computer Science and Technology, University of Chinese Academy of Sciences, Beijing 100040, China (e-mail: tfan.108@gmail.com).}
}
% \markboth{IEEE Transactions on Image Processing,~Vol.~XX, No.~XX, 2025}%


\maketitle

\begin{abstract}

Style transfer, a pivotal task in image processing, synthesizes visually compelling images by seamlessly blending realistic content with artistic styles, enabling applications in photo editing and creative design. While mainstream training-free diffusion-based methods have greatly advanced style transfer in recent years, their reliance on computationally inversion processes compromises efficiency and introduces visual distortions when inversion is inaccurate.
To address these limitations, we propose a novel \textit{inversion-free} style transfer framework based on dual rectified flows, which tackles the challenge of finding an unknown stylized distribution from two distinct inputs (content and style images), \textit{only with forward pass}. Our approach predicts content and style trajectories in parallel, then fuses them through a dynamic midpoint interpolation that integrates velocities from both paths while adapting to the evolving stylized image.
By jointly modeling the content, style, and stylized distributions, our velocity field design achieves robust fusion and avoids the shortcomings of naive overlays. Attention injection further guides style integration, enhancing visual fidelity, content preservation, and computational efficiency. Extensive experiments demonstrate generalization across diverse styles and content, providing an effective and efficient pipeline for style transfer.
The code will be available at \href{https://github.com/HolmesShuan/Inversion-Free-Style-Transfer-with-Dual-Rectified-Flows}{this-URL}.
\end{abstract}


% \IEEEPARstart{T}{his} file is intended to serve as a ``sample article file''
% for IEEE journal papers produced under \LaTeX\ using
% IEEEtran.cls version 1.8b and later. The most common elements are covered in the simplified and updated instructions in ``New\_IEEEtran\_how-to.pdf''. For less common elements you can refer back to the original ``IEEEtran\_HOWTO.pdf''. It is assumed that the reader has a basic working knowledge of \LaTeX. Those who are new to \LaTeX \ are encouraged to read Tobias Oetiker's ``The Not So Short Introduction to \LaTeX ,'' available at: \url{http://tug.ctan.org/info/lshort/english/lshort.pdf} which provides an overview of working with \LaTeX.
\section{Introduction}
\IEEEPARstart{I}{mage} stylization allows users to create visually striking compositions by blending realistic content with artistic styles. This process has wide applications in digital art, image editing, and advertising, enabling professional-quality results without advanced skills and making creative tools more accessible.

\begin{figure}
\centering
\includegraphics[width=1.0\linewidth]{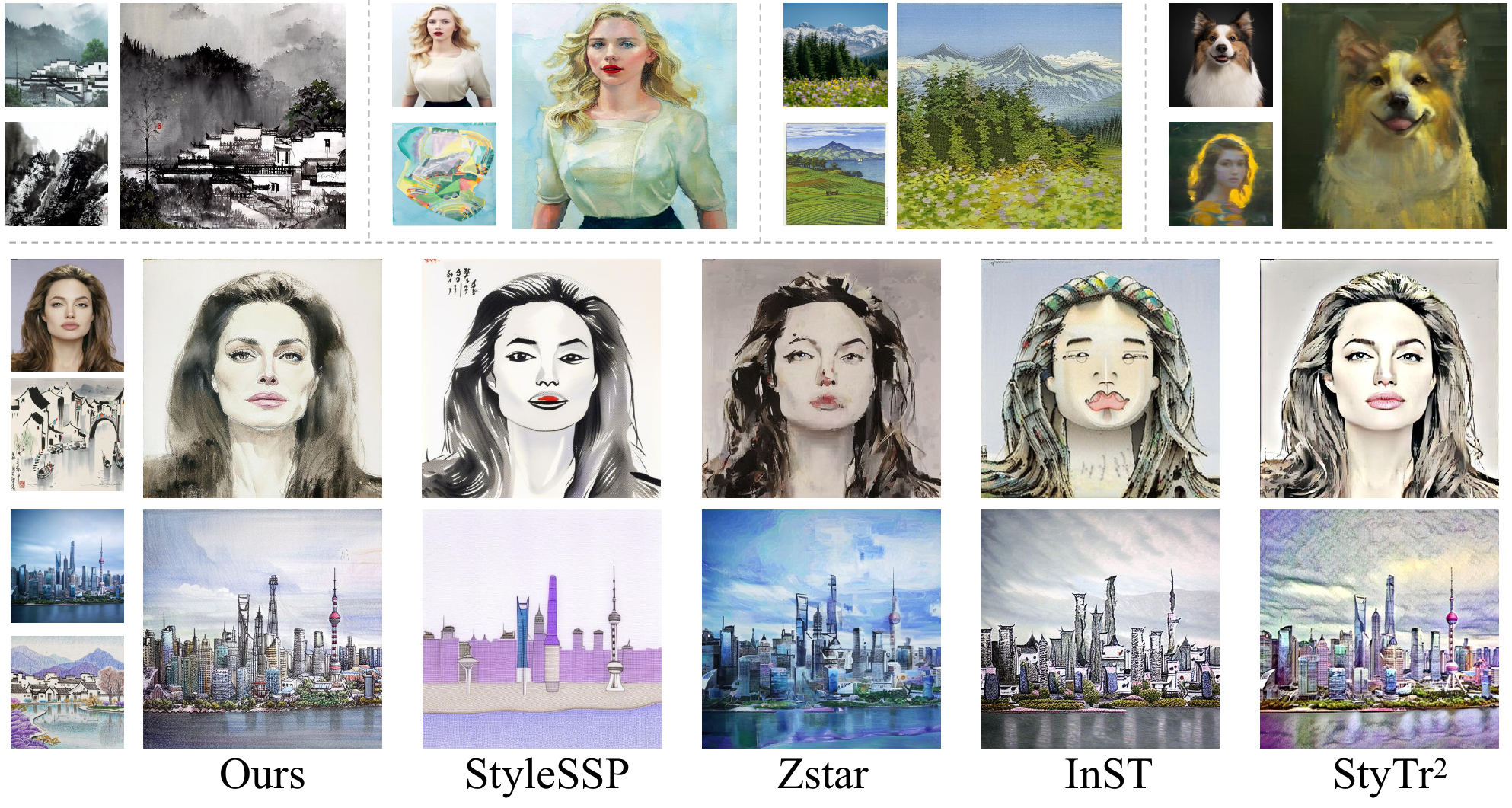}
\caption{Image style transfer results from our proposed method.
\textbf{First row}: Stylized outputs using various style and content references.
\textbf{Last two rows}: Comparisons with state-of-the-art methods, including diffusion-based style transfer (inversion-based training-free models like StyleSSP~\cite{StyleSSP} and Zstar~\cite{Deng_2024_CVPR}; fine-tuning-based InST~\cite{zhang:2023:inversion}) and traditional StyTr$^2$~\cite{Deng:2022:CVPR}. Our approach preserves core structural details while delivering vibrant, distinctive artistic flair.
}
\label{fig:f1}
\end{figure}

Generative modeling has advanced style transfer by offering strong representations of content and style. Diffusion models, in particular, have driven many state-of-the-art methods~\cite{Deng_2024_CVPR} due to their powerful generation abilities. A typical approach uses pre-trained text-to-image models, extracting styles via fine-tuning~\cite{zhang:2023:inversion, dreambooth}. However, these methods often rely on explicit, sequential separation of content and style during generation, leading to high computational costs and incomplete disentanglement. This can alter content or weaken styles (see InST results in Fig.~\ref{fig:f1}).
Training-free methods~\cite{Deng_2024_CVPR} reduce these costs but depend on precise latent features from DDIM (Denoising Diffusion Implicit Models) inversion. In practice, even minor inversion inaccuracies can propagate through the generation chain, causing content distortion and loss of coherence (see StyleSSP and Zstar results in Fig.~\ref{fig:f1}). Rectified flow (ReFlow) models~\cite{labs2025flux,EsserKBEMSLLSBP24,opensora2025} have recently emerged as a better alternative to diffusion models, with faster inference, higher fidelity, and broad use in high-resolution synthesis. ReFlow's straight-line trajectories produce superior results, ideal for image editing. Yet many top editing methods~\cite{wang2025taming,FireFlow, Multi-turn} still use inversion, adding complexity and risking distortions. In contrast, FlowEdit~\cite{flowedit2025} offers an inversion-free option for editing, avoiding these issues while maintaining quality.

While inversion-free methods work well for single-trajectory editing, style transfer is more complex, involving two inputs (content and style) and dual trajectories. Unlike single-trajectory tasks, where inversion-free methods can directly map a noisy image to a modified target, style transfer demands the construction of an unknown stylized distribution that dynamically integrates elements from both content and style inputs. Naively extending existing inversion-free techniques risks reducing the process to simplistic overlays of content and style, resulting in poor stylization quality that fails to preserve structural integrity or capture artistic details. Furthermore, while inversion-free designs eliminate reconstruction errors associated with inversion, style transfer requires more than error minimization. It necessitates a sophisticated velocity field construction that simultaneously accounts for the content image, style image, and evolving stylized result to guide the trajectory toward a coherent and visually compelling output. Without carefully designed fusion mechanisms, the process struggles to balance content preservation and style infusion, often leading to artifacts or diluted artistic effects.

To address these challenges and unlock the full potential of inversion-free style transfer, we introduce a dual rectified flows framework that relies solely on forward ODE processes. Built on pre-trained ReFlow models, our approach predicts content and style trajectories in parallel, fusing them through a novel midpoint interpolation heuristic that dynamically integrates the velocities from both paths while accounting for the evolving stylized image. Constructing the velocity field by simultaneously considering the content, style, and stylized distributions ensures robust fusion without degenerating into simple additions. Furthermore, we incorporate attention injection into the forward process to more effectively guide the trajectory direction, enhancing style cue integration. This unified pipeline tackles the dual challenges of computational simplicity and high-quality stylization, delivering enhanced stability, reduced memory usage, and generalization across diverse artistic styles and content types.
In summary, our main contributions are as follows:
\begin{itemize}
\item We tackle the instability and cost of inversion in training-free style transfer methods by leveraging forward ODE trajectories for efficient, high-fidelity stylization.
\item We propose a novel dual rectified flows framework that parallelizes content and style trajectories, fusing them through midpoint interpolation to dynamically build velocities considering content, style, and evolving stylized images.
\item We integrate attention injection for better style guidance and conduct comprehensive experiments demonstrating superior visual fidelity, content preservation, and generalization across diverse artistic styles compared to SOTA baselines.
\end{itemize}

\IEEEpubidadjcol
\section{Related Work}
\subsection{Traditional Style Transfer}

The traditional image style transfer tasks have been extensively researched using diverse network architectures, including Convolutional Neural Networks (CNNs)~\cite{gatys:2016:image, Huang:2017:Arbitrary,li:2017:universal,Semantic,2012tip,2022tip}, 
Transformers~\cite{Deng:2022:CVPR}, flow ~\cite{an:2021:artflow}, and Mamba-based networks~\cite{2025cvprsamam}. These methods typically rely on predefined style and content losses to guide the stylization process. Gatys et al.~\cite{gatys:2016:image} first demonstrate that the Gram matrix, computed as the inner product of feature maps across channels, effectively captures style information. Leveraging this insight, subsequent works~\cite{johnson:2016:perceptual,li:2016:precomputed} developed end-to-end models to enable faster style transfer.
Li et al.~\cite{johnson:2016:perceptual} propose replacing per-pixel losses with perceptual losses to train feed-forward networks for image transformation tasks, thereby achieving real-time style transfer.
Li et al.~\cite{li:2016:precomputed} propose a feed-forward convolutional generator trained via adversarial learning on Markovian patch statistics, enabling real-time texture synthesis.

To facilitate better fusion of content and style features, WCT~\cite{li:2017:universal} introduces a whitening and coloring transform module. AdaIN~\cite{Huang:2017:Arbitrary} proposes adaptive instance normalization for dynamic style adaptation. SANet~\cite{park:2019:arbitrary} incorporates a style-attention mechanism to enable local content-style matching. More recently, HSI~\cite{2025cvprHSI} proposes an attention-based style transformation module to enhance the artistic expressiveness of the target style.
In addition to CNN-based approaches, ArtFlow~\cite{an:2021:artflow} employs a flow-based model to address content leakage issues arising from CNN structures. StyTr$^2$~\cite{Deng:2022:CVPR} introduces the first Transformer-based model to achieve superior stylization performance through global context modeling. SaMam~\cite{2025cvprsamam} utilizes a Mamba-based architecture to balance global receptive fields with linear computational complexity. However, such global style representations often compromise fine-grained detail preservation in stylized outputs.
To mitigate this limitation, CAST~\cite{zhang:2022:domain} and CLAST~\cite{CLAST} leverage contrastive learning to impose local style constraints. Similarly, CCPL~\cite{2022contrastive} encourages consistency between the content image and generated image in terms of the difference of an image patch with its neighboring patches.

Despite the continuous advancements in existing methods, achieving precise style representation remains a formidable challenge, as inaccuracies in style modeling can result in suboptimal stylized outcomes. In this work, we delve into the style and content trajectories of pre-trained text-to-image (T2I) models to enable training-free style transfer.

\subsection{T2I Models for Style Transfer}
Recent advancements in T2I models have significantly influenced style transfer techniques, enabling more controllable and semantically rich stylization through diffusion-based or flow-based architectures~\cite{TRTST,Deng_2024_CVPR,zhang:2023:inversion,CSGO}.

InST~\cite{zhang:2023:inversion} and VCT~\cite{cheng:2023:ict} employ an inversion-based image style transfer/translation scheme that can train a style image into a style embedding to guide the generated results. 
B-LoRA~\cite{b-lora} introduces jointly learning LoRA for implicit style-content separation using dual-branch Low-Rank Adaptation.
B4M ~\cite{B4M} deconstructs and re-purposes Low-Rank Adapters to achieve independent control over content and style in image generation.
DEADiff~\cite{DEADiff} proposes a dual-branch architecture that extracts features through separate learning pathways, isolating stylistic attributes from semantic content to prevent interference.
CSGO~\cite{CSGO} achieves fine-grained control in text-to-image generation by separately encoding and composing content and style representations, enabling the flexible creation of images with specified semantic content and artistic style.
InstantStyle~\cite{InstantStyle} fine-tunes diffusion models with style-preserving losses to transfer reference styles without heavy optimization.
ArtAdapter~\cite{ArtAdapter} employs multi-level adaptation modules fine-tuned on style datasets to disentangle and transfer artistic elements in T2I pipelines, outperforming baselines in expressive stylization.

However, the dataset lack and training consumption limit the generalization ability of style transfer methods. Therefore, some researchers focus on training-free methods that primarily rely on feature inversion in the latent space to fuse content and style representations.
StyleAligned~\cite{Style-Aligned} enforces style consistency through minimal attention sharing and inversion, enabling high-quality synthesis across diverse prompts without fine-tuning.
Deng et al.~\cite{Deng_2024_CVPR,dengz+} rearrange attention maps in the latent space without fine-tuning, directly extracting and integrating style priors from vanilla diffusion models into content latents for seamless transfer.
StyleID~\cite{styleid} replaces the Key and Value features in the self-attention layers of the content image with those from the style image without requiring any fine-tuning or optimization for each new style.
Building upon cross-attention rearrangement, some studies further aim to enhance the quality of generated images.
StyleKeeper~\cite{stylekeep} addresses content leakage in visual style prompting for text-to-image diffusion models by extending classifier-free guidance with self-attention swapping and introducing negative visual query guidance, which simulates leakage via query swapping to minimize unwanted content transfer while preserving desired styles.
StyleSSP~\cite{StyleSSP} enhances the starting point of the sampling process in a diffusion model to improve its performance in style transfer tasks.

Despite these innovations, current T2I-based style transfer methods face several critical challenges that compromise their practical utility. A primary issue is the heavy reliance on inversion processes, such as DDIM inversion, to map content images into latent spaces for denoising-based stylization, which often introduces computational overhead and stochastic distortions, leading to content leakage or geometric inconsistencies. In this work, we introduce a dual rectified flows framework that eliminates compromises in efficiency and visual distortions through an inversion-free approach.

\begin{figure*}
\centering
\includegraphics[width=1.0\linewidth]{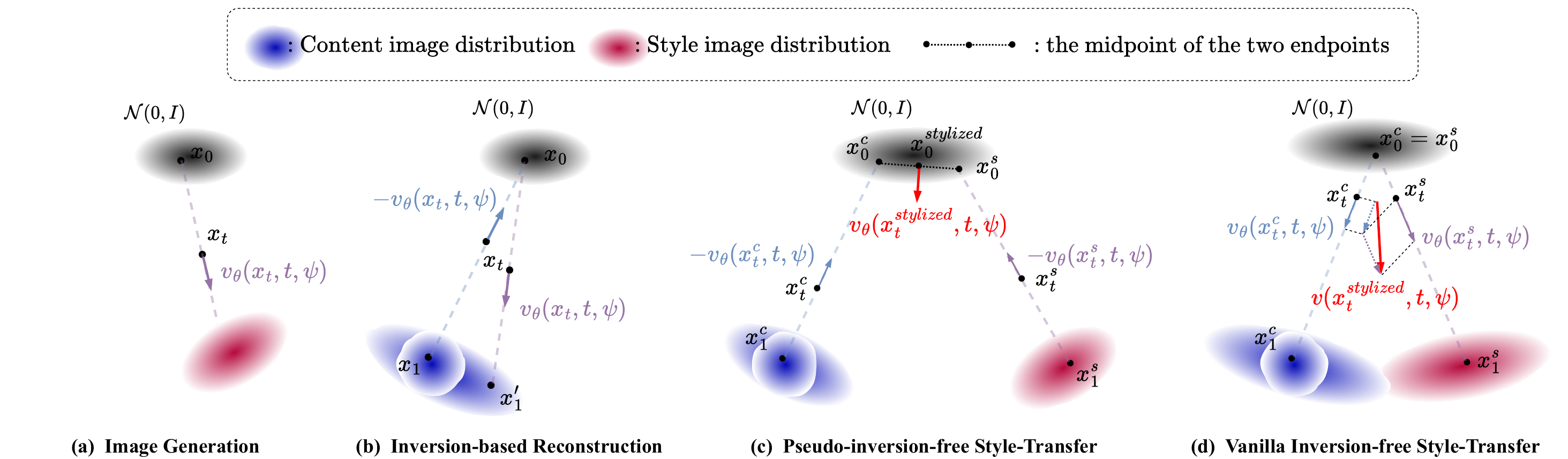}
\caption{From image generation to style-transfer:
(a) Image generation via the ReFlow model begins by sampling random noise \(x_0\) from a Gaussian distribution. A velocity field \(v_\theta(x, t, \psi)\), which depends on a text prompt \(\psi\), is then employed to generate the image \(x_1\).
(b) To obtain the proper noise \(x_0\) that can better reconstruct the content image \(x_1\), we first solve the ODE conditioned on the source prompt \(\psi\) to find the appropriate noise \(x_0\). We then follow the image generation path to obtain the reconstructed image \(x_1'\).
(c) Assuming that the oracle content noise \(x_0^c\) and style noise \(x_0^s\) are known via inversion, we can interpret the pseudo noise \(x_0^{stylized}\) using \(x_0^c\) and \(x_0^s\). It is then straightforward to use the velocities predicted by \(x_t^{stylized}\) to form a new direction \(v_\theta(x_t^{stylized}, t, \psi)\) towards stylized image.
(d) To avoid the inversion process, we directly add random noise to \(x_1^c\) and \(x_1^s\). The noisy image \(tx_1 + (1-t)x_0\) shows the same starting point \(x_0^c = x_0^s\) for the forward ODEs. Then, by utilizing vector addition, we use the directions of content and style image denoising to form the new velocity \(v(x_t^{{stylized}}, t, \psi)\) for style transfer.}
\label{fig:framework_1}
\end{figure*}

\section{Preliminaries}
In this section, we start with a brief introduction to the formulation of the Rectified Flow model. We then illustrate how to perform basic image generation and image reconstruction, which aids in understanding the motivation behind our approach.

\subsection{Rectified Flow}

Rectified Flow \cite{LiuG023} provides a principled framework for transforming between two distributions $\pi_0$ and $\pi_1$ using paired observations $(x_0, x_1) \sim \pi_0 \times \pi_1$. The method models the transformation via an ordinary differential equation (ODE) over time $t \in [0, 1]$:
\begin{equation}
d x_t = v(x_t, t)  dt,
\label{eq:ode}
\end{equation}
where $x_0 \sim \pi_0$ and $x_1$ follows $\pi_1$. The drift function $v: \mathbb{R}^d \times [0, 1] \to \mathbb{R}^d$ is learned to align the ODE trajectory with the straight path connecting $x_0$ and $x_1$, by solving the regression problem:
\begin{equation}
\min_v\ \mathbb{E} \left[ \int_0^1 \left|| (x_1 - x_0) - v_\theta(x_t, t) \right||_2^2 dt \right],
\end{equation}
with $x_t = t x_1 + (1 - t) x_0$ representing the linear interpolation.

\noindent\textbf{Forward Process.} The goal is to transport samples from $\pi_0$ toward $\pi_1$. Although the linear interpolation $x_t = t x_1 + (1 - t) x_0$ satisfies the ODE
\begin{equation}
dx_t = (x_1 - x_0)  dt.
\end{equation}
This formulation is non-causal as it requires knowing $x_1$ in advance. Rectified Flow addresses this by introducing a learned drift $v(x_t, t)$ that approximates the direction $x_1 - x_0$, yielding a causal forward process:
\begin{equation}
dx_t = v(x_t, t)  dt, \quad x_0 \sim \pi_0,\ t\in[0,1].
\label{eq:ode_forward}
\end{equation}
This allows the simulation of $x_t$ without access to the endpoint $x_1$ during integration.

\noindent\textbf{Reverse Process.} To generate samples from $\pi_1$, the forward ODE is reversed in time. Starting from $x_1 \sim \pi_1$, the reverse dynamics are given by:
\begin{equation}
dx_t = -v(x_t, t)  dt, \quad t \in [1, 0].
\label{eq:ode_backward}
\end{equation}
This backward process inverts the forward transformation, effectively mapping samples from $\pi_1$ back to $\pi_0$ while maintaining consistency through the symmetry of the learned drift.

\subsection{Image Generation via Rectified Flow}
Given a pre-trained Rectified Flow model, image generation involves sampling from the target distribution $\pi_1$ (e.g., the distribution of images) by solving the forward ODE learned during training. The model transports samples from a source distribution $\pi_0$ (e.g., Gaussian noise $\mathcal{N}(0, I)$) to $\pi_1$ using a neural network-parameterized drift function ($v_\theta: \mathbb{R}^d \times [0, 1] \to \mathbb{R}^d$) and prompt $\psi$, optimized to approximate straight-line trajectories.

\noindent\textbf{Image Generation.} To generate an image, initialize a noise sample $x_0 \sim \pi_0$, where $x_0$ has the same dimensionality as the target images (e.g., $3 \times 512 \times 512$ for RGB images). The forward ODE is solved numerically:
\begin{equation} 
dx_t = v_\theta(x_t, t, \psi) dt, \quad x_0 \sim \pi_0, \quad t \in [0, 1]. \label{eq:ode_sampling} 
\end{equation}
to obtain $x_1 \sim \pi_1$, the generated image. The integration is typically performed using a simple Euler method for efficiency:
\begin{equation} 
x_{(i+1)/N} = x_{i/N} + \frac{1}{N} v_\theta(x_{i/N}, i/N), \end{equation}
where $N$ is the number of discrete steps. After integration, $x_1$ represents the generated image, which may require post-processing (e.g., pixel normalization to $[0, 255]$). The whole process is shown in Fig. \ref{fig:framework_1}(a).

\noindent\textbf{Inversion-based Reconstruction.} For the tasks requiring content preservation, such as style transfer, inversion-based image reconstruction exploits the invertibility of the learned ODE to map a source image back to its latent representation (typically noise) and then reconstruct it through the forward process. As shown in Fig. \ref{fig:framework_1}(b), given a source image $x_1 \sim \pi_1$, the inversion computes the corresponding latent noise $x_0 \sim \mathcal{N}(0, I)$ by integrating the reverse ODE:
\begin{equation} 
dx_t = -v_\theta(x_t, t, \psi) dt, \quad x_1 \sim \pi_1, \quad t \in [1, 0], \label{eq:ode_inversion} 
\end{equation}
where $v_\theta$ is the pre-trained drift function. Numerical integration (e.g., via Euler method) yields an approximate $\hat{x}_0$, capturing the image's latent encoding. However, due to discretization errors or imperfect training, this step commonly introduces distortions. To reconstruct the image, the forward ODE is applied to the inverted latent $\hat{Z}_0$:
\begin{equation} 
dx_t = v_\theta(x_t, t, \psi) dt, \quad x_0 \sim \mathcal{N}(0, I), \quad t \in [0, 1], \label{eq:ode_reconstruction} 
\end{equation}
resulting in a reconstructed $x'_1 \approx x_1$. The quality of reconstruction heavily depends on the ODE's invertibility and the number of integration steps $N$.

\section{Approach}
In this section, we show our attempts to solve the inversion-free problem step by step, from trivial solutions to our final solution. At first, we explore a trivial solution to achieving pseudo-inversion-free style transfer by mimicking the inversion and reconstruction processes found in the vanilla ReFlow model. Subsequently, we will explore methods to eliminate the necessity of inversion and introduce our dual rectified flows model, which enhances the utilization of predicted velocity from both content and style images during the solution of ordinary differential equations (ODEs).

\subsection{Pseudo-inversion-free Style-Transfer}
Following the inversion-based image reconstruction pipeline, it is intuitive to derive the pseudo-inversion-free setting \footnote{The term "pseudo-inversion-free" indicates that, while the stylized latent noise $x_0^{stylized}$ is generated without directly inverting a stylized image, the process still relies on inverting the source content and style images to obtain $x_0^c$ and $x_0^s$, respectively. Thus, it is not truly inversion-free.} by leveraging the latent noises $x_0^c$ and $x_0^s$, obtained through inversion of the content and style images shown in Figure \ref{fig:framework_1}(c). The stylized latent noise $x_0^{stylized}$ is constructed as a function of $x_0^c$ and $x_0^s$, typically by combining or interpolating these latent states to encode both content and style properties, formally expressed as $x_0^{stylized} = f(x_0^c, x_0^s)$.

With $x_0^{stylized}$ defined, the stylized image is synthesized using the forward ordinary differential equation (ODE): 
\begin{equation} dx_t = v_\theta(x_t^{stylized}, t, \psi) dt, \quad x_0^{stylized}, \quad t \in [0, 1], \label{eq:ode_stylized} 
\end{equation} 
where $v_\theta$ denotes the pre-trained drift function. Integrating this ODE yields a stylized image that retains the content structure of $x_0^c$ while adopting the style features of $x_0^s$, guided by the predicted velocities for $x_t^{stylized}$. Despite its intuitive derivation, a key limitation of this approach, assuming the correctness of $x_0^{stylized}$, is its continued reliance on inversion to generate $x_0^c$ and $x_0^s$.

\subsection{Vanilla Inversion-free Style-Transfer}
\label{sec:vanilla_inv_free}
To further eliminate the need for any inversion steps, we start from a direct approach. Instead of computing latent noise representations by inverting source images, we return to the definition of rectified flow, where the ODE trajectory shows the straight path connecting \(x_0\) and \(x_1\) by simply sampling random noise vectors and adding them directly to their respective images, \(x_1^c\) (content) and \(x_1^s\) (style). These noisy images, \(tx_1^c + (1-t)x_0^c\) and \(tx_1^s + (1-t)x_0^s\), describe the initial conditions \(x_0^c = x_0^s\) for the forward ODEs.

During the forward process, the denoising directions for both content and style images are obtained by integrating the ODEs starting from their respective noisy initializations. At each time step \(t\), the velocities (i.e., denoising directions) predicted for the content and style images are combined—typically via vector addition—to construct the velocity for the stylized image:
\begin{align}
v(x_t^{{stylized}}, t, \psi) = \lambda \cdot v_\theta(x_t^c, t, \psi) + (1-\lambda) \cdot v_\theta(x_t^s, t, \psi). \label{eq:ode_vanilla_stylized}
\end{align}
Though this approach entirely avoids the inversion process, as no latent codes are computed by reversing the ODE, it is easy to prove that the denoised stylized image is indeed the mean value of the content and style images if the reflow model is well-trained with \(\lambda = \frac{1}{2}\):
\begin{equation}
\begin{aligned}
    x_1^{{stylized}} &= x_0^{{stylized}} + v(x_t^{{stylized}}, t, \psi) \\
    &= x_0^c + \frac{1}{2}(x_1^c - x_0^c) + \frac{1}{2}(x_1^s - x_0^s) = \frac{1}{2}(x_1^c + x_1^s),
\end{aligned}
\end{equation}
since \(v_\theta(x_t, t, \psi) = x_1 - x_0\). It is obvious that this trivial solution avoids inversion at the cost of losing accurate content preservation or stylization, which degrades into the simple addition of content and style images.

\begin{figure*}
\centering
\includegraphics[width=1.0\linewidth]{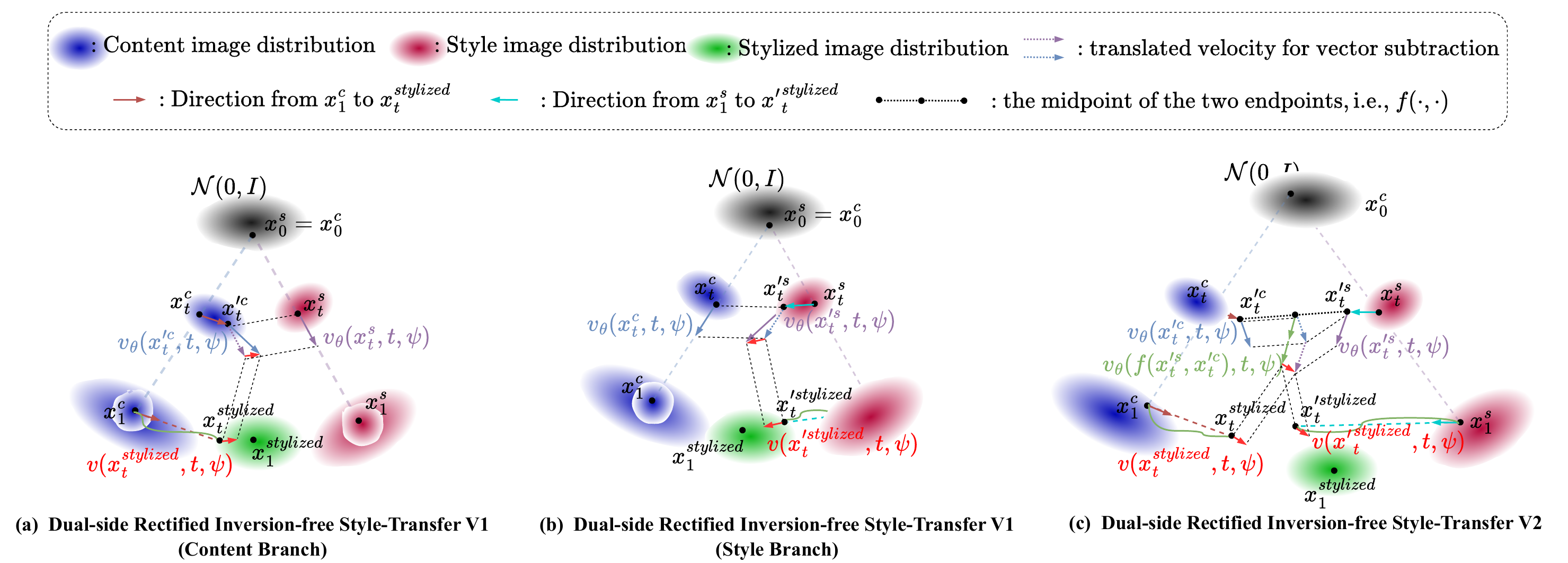}
\caption{Illustration of our style transfer trajectories in rectified flow space. a) The content trajectory (blue dotted line) starts from the shared Gaussian noise \( x_0^c \sim \mathcal{N}(0,1) \) at \( t=0 \) and moves toward the content image \( x^c_1 \) at \( t=1 \). The style trajectory (purple dotted line) begins at the same noise \( x_0^s = x_0^c \), converging to the style image \( x^s_1 \) at \( t=1 \). The shifted content trajectory (dark red arrow) adjusts the content path by \( \tau (x^{stylized}_t - x^c_1) \), gradually aligning with the evolving stylized image \( x^{stylized}_t \). The velocity of \( x^{stylized}_t \) (red arrow) is formulated as the vector subtraction of the style image $x_t^s$'s velocity and the shifted content image $x'^c_t$'s velocity, targeting the style image distribution. These straight-line paths couple content and style via shared noise, guiding the transformation from the content image to the final stylized image at \( t=1 \). (b) Similar to the content-branch generation, it is straightforward to derive the mirrored style-branch inversion-free style transfer, where we use the style image as the initial point moving toward the content image. (c) We aim to balance content and style by deriving from both the content and style branches; hence, the midpoint was introduced as a natural combination of \( x'^c_t \) and \( x'^s_t \), predicting the fixed velocity $v_\theta(f(x'^s_t,x'^c_t), t, \psi)$, which is then refined by the velocities of the shifted content and style images (\( v_\theta(x'^c_t, t, \psi) \) and \( v_\theta(x'^s_t, t, \psi) \)) to form the final velocity (red arrow).}
\label{fig:framework_3}
\end{figure*}

\subsection{Dual Rectified Flows Inversion-free Style-Transfer}
\subsubsection{Version I}
As the simple addition of velocities starting from random noise fails to generate meaningful results (as shown in Section \ref{sec:vanilla_inv_free}), we instead adopt velocity subtraction. This approach establishes a straight path from the content image to the style image (and vice versa), enabling a direct and intuitive fusion of content and style.

Formally, we define the velocity for the stylized image as
\begin{equation}
    \label{eq:velocity_0}
    v(x_t^{stylized}, t, \psi) = v_\theta(x_t^s, t, \psi) - v_\theta(x_t^c, t, \psi),
\end{equation}
where \( v_\theta \) is the pre-trained velocity field. This velocity corresponds to a path starting from the content image \( x_1^c \) and moving toward the difference \( x_t^s - x_t^c \):
\begin{equation}
    x_t^{stylized} = x_1^c + (x_t^s - x_t^c).
    \label{eq:path1}
\end{equation}
To derive the corresponding ODE, differentiate Equation~\eqref{eq:path1} with respect to \( t \):
\begin{align}
    \label{eq:velocity_1}
    dx_t^{stylized} = dx_t^s - dx_t^c &= v_\theta(x_t^s, t, \psi) \, dt - v_\theta(x_t^c, t, \psi) \, dt.\\
    \label{eq:velocity_1_1}
    &=\big(v_\theta(x_t^s, t, \psi) - v_\theta(x_t^c, t, \psi)\big)dt
\end{align}
This yields the ODE for the stylized trajectory, satisfying the boundary conditions \( x_0^{stylized} = x_1^c \) and \( x_1^{stylized} = x_1^s \).

Our earlier attempts in Section \ref{sec:vanilla_inv_free} simply combined the velocities of the known content \( x_t^c \) and style \( x_t^s \) trajectories to form \( v(x_t^{stylized}) \). However, this neglects the evolving properties of the stylized image $x_t^{stylized}$ itself, hence degrades into a trivial solution. Equation (\ref{eq:velocity_1_1}) faces the same problem that no $x_t^{stylized}$ involved in the prediction of velocity. To address this, we reformulate Equation~\eqref{eq:velocity_1} from the perspective of \( x_t^{stylized} \), expressing the relationship between \( x_t^c \) and \( x_t^s \) in terms of the stylized path. Rearranging Equation~\eqref{eq:path1} gives:
\begin{equation}
    x_t^s = x_t^c + (x_t^{stylized} - x_1^c).
    \label{eq:path2}
\end{equation}
Substituting this into Equation~\eqref{eq:velocity_1}, we obtain:
\begin{equation}
    \label{eq:velocity_2}
    dx_t^{stylized} = v_\theta(x_t^{stylized} + x_t^c - x_1^c, t, \psi) \, dt - v_\theta(x_t^c, t, \psi) \, dt.
\end{equation}

Symmetrically, we define a mirrored velocity for the path starting from the style image toward the content image:
\begin{equation}
    v({x'}_t^{stylized}, t, \psi) = v_\theta(x_t^c, t, \psi) - v_\theta(x_t^s, t, \psi),
    \label{eq:velocity_4}
\end{equation}
with the corresponding path:
\begin{equation}
\begin{aligned}
    {x'}_t^{stylized} &= x_1^s + (x_t^c - x_t^s), \\
    \label{eq:path3}
    x_t^c &= {x'}_t^{stylized} + x_t^s - x_1^s.
\end{aligned}
\end{equation}
Differentiating and substituting yields the ODE:
\begin{equation}
    \label{eq:velocity_3}
    d{x'}_t^{stylized} = v_\theta({x'}_t^{stylized} + x_t^s - x_1^s, t, \psi) \, dt - v_\theta(x_t^s, t, \psi) \, dt.
\end{equation}

Although Equations~\eqref{eq:velocity_2} and \eqref{eq:velocity_3} describe stylized trajectories that incorporate \( x_t^{stylized} \) (or \( {x'}_t^{stylized} \)) into the velocity prediction at timestep \( t \), they depend solely on either the content trajectory \( x_t^c \) or the style trajectory \( x_t^s \). This limitation prevents full access to both content and style information simultaneously, potentially leading to incomplete stylization or content preservation.

To overcome this, we combine the two velocities \( v(x_t^{stylized}, t, \psi) \) and \( v({x'}_t^{stylized}, t, \psi) \), which together consider \( x_t^s \), \( x_t^c \), and the evolving stylized image. Adding Equations~\eqref{eq:velocity_2} and \eqref{eq:velocity_3} gives:
\begin{equation}
\begin{aligned}
    &dx_t^{stylized} + d{x'}_t^{stylized} \\
    &= v(x_t^{stylized}, t, \psi) \, dt + v({x'}_t^{stylized}, t, \psi) \, dt \\
    &= \big[ v_\theta(x_t^{stylized} + x_t^c - x_1^c, t, \psi) - v_\theta(x_t^c, t, \psi) \big] \, dt \\
    &\quad + \big[ v_\theta({x'}_t^{stylized} + x_t^s - x_1^s, t, \psi) - v_\theta(x_t^s, t, \psi) \big] \, dt \\
    &= \underbrace{\big( v_\theta(x_t^{stylized} + x_t^c - x_1^c, t, \psi) - v_\theta(x_t^s, t, \psi) \big) \, dt}_{\text{Content-branch Rectified Inversion-free Style-transfer}} \\
    &\quad + \underbrace{\big( v_\theta({x'}_t^{stylized} + x_t^s - x_1^s, t, \psi) - v_\theta(x_t^c, t, \psi) \big) \, dt}_{\text{Style-branch Rectified Inversion-free Style-transfer}}\\
    &=2\cdot d(\frac{1}{2}x_t^{stylized} + \frac{1}{2}{x'}_t^{stylized}).
\end{aligned}
\end{equation}
This summation approximates the velocity at the midpoint between \( x_t^{stylized} \) and \( {x'}_t^{stylized} \), effectively balancing both paths (transfer from content images to style images and transfer from style images to content images) for a more robust stylized result. To better visually illustrate these ODE trajectories, we decompose the process into two sub-figures, corresponding to Fig. \ref{fig:framework_3}(a) and Fig. \ref{fig:framework_3}(b), which highlight the predicted velocities in each branch. The complete dual-side process is detailed in Algorithm \ref{alg:content-style-side}.

\subsubsection{Version II} 

While our content-branch and style-branch rectified inversion-free velocities (Equations~\eqref{eq:velocity_2} and \eqref{eq:velocity_3}) effectively initiate style transfer by guiding the trajectory from content to style (and vice versa), they exhibit limitations in capturing the full interplay of content and style, particularly at later timesteps. For instance, consider the style-side velocity:
\begin{equation}
\begin{aligned}
    \lim_{t\to 1} \big[ v_\theta({x'}_t^{stylized} + &x_t^s - x_1^s, t, \psi) - v_\theta(x_t^c, t, \psi) \big] \\
    &\approx v_\theta({x'}_t^{stylized}, t, \psi) - v_\theta(x_t^c, t, \psi).
\end{aligned}
\label{eq:velocity_dual_side_v2_vanilla}
\end{equation}
As \( t \to 1 \), the term \( {x'}_t^{stylized} + x_t^s - x_1^s \) approaches \( {x'}_t^{stylized} \) contributing to the ideal velocity $v_\theta({x'}_t^{stylized}, t, \psi)$. Meanwhile, the second term velocity $v_\theta(x_t^c, t, \psi)$ becomes gradually independent of the current stylized image \( x_1^{stylized} \). Similarly, the content-branch velocity \( v_\theta(x_t^{stylized} + x_t^c - x_1^c, t, \psi) - v_\theta(x_t^s, t, \psi) \) converges to \( v_\theta(x_t^{stylized}, t, \psi) - v_\theta(x_t^s, t, \psi) \), where the second term losing connections with the stylized image \( x^{stylized}_t \). Early in the process (low \( t \)), when \( x_t^{stylized} \) is noisy and undefined, the content and style trajectories provide essential coarse guidance. However, at later stages (high \( t \)), the differences between $x^{stylized}_t$ and $x_t^c$ (or $x_t^s$) becomes larger, which hinders fine-grained stylization, leading to suboptimal content preservation or style fidelity.

To address this, we propose a heuristic that integrates both content and style information throughout the trajectory by interpolating the content-branch and style-branch paths. Inspired by an artist’s iterative refinement, where content and style are balanced dynamically, we introduce a midpoint trajectory \( x'_t \) defined as:
\begin{equation}
    x'_t = \frac{1}{2} (x_t^c + x_t^{stylized} - x_1^c) + \frac{1}{2} (x_t^s + {x'}_t^{stylized} - x_1^s).
    \label{eq:midpoint}
\end{equation}
This interpolated point captures the essence of both trajectories, with limiting behaviors:
\begin{equation}
\begin{aligned}
    \lim_{t \to 0} x'_t &\approx \frac{1}{2} (x_0^c + x_0^s), \\
    \lim_{t \to 1} x'_t &\approx \frac{1}{2} (x_t^{stylized} + {x'}_t^{stylized}).
    \label{eq:midpoint_limits}
\end{aligned}
\end{equation}
At early timesteps (\( t \to 0 \)), \( {x'}_t \) leverages the shared noise of \( x_0^c \) and \( x_0^s \), providing robust initial guidance akin to the original trajectories. At later timesteps (\( t \to 1 \)), it approximates the midpoint between the content-branch and style-branch stylized estimates, ensuring a balanced fusion that reflects both content structure and style aesthetics.

By incorporating \( {x'}_t \), we redefine the velocity to simultaneously account for \( x_t^c \), \( x_t^s \), \( x_t^{stylized} \), and \( {x'}_t^{stylized} \) to replace the second term in Equation (\ref{eq:velocity_dual_side_v2_vanilla}). This velocity drives a new ODE that combines the content-branch and style-branch contributions, as shown in Algorithm~\ref{alg:dual-side}. The process is visualized in Fig.~\ref{fig:framework_3}(c). This approach ensures a seamless, artist-inspired synthesis, enhancing visual coherence and robustness across diverse styles, as demonstrated in our experiments.

\begin{algorithm}[tb]
   \caption{Dual-side Rectified Inversion-free Style-Transfer V1}
   \label{alg:content-style-side}
\begin{algorithmic}
   \STATE {\bfseries Input:} 
   content image $x_1^c$, style image $x_1^s,\big\{t_i\big\}^{n_{\text{max}}}_{i=0}, n_{\text{max}},$ scaling factor $\tau=1$ 
   \STATE {\bfseries Output:} stylized image $x^{stylized}_1$
   \STATE {\bfseries Init:} $x^{stylized}_{t_0}=x^{c}_1,\quad x'^{stylized}_{t_0}=x^{s}_1$
   \FOR{$i=0$ {\bfseries to} $n_{\text{max}}$}
   \STATE $x_0^s \leftarrow x_0^c,\quad x_{0}^s \sim \mathcal{N}(0,\,1)$
   \STATE $x^{c}_{t_i} \leftarrow (1-t_i)x^c_1 + t_ix_0^c,\quad x^{s}_{t_i} \leftarrow (1-t_i)x^s_1 + t_ix_0^s$
   \STATE $x'^{c}_{t_i} \leftarrow x^{c}_{t_i} +\tau(x^{stylized}_{t_i}- x^{c}_1)$
   \STATE $v(x_{t_i}^{stylized},t_i,\psi) \leftarrow -v_\theta(x^{s}_{t_i},t_i,\psi)+v_\theta(x'^{c}_{t_i},t_i,\psi)$
   \STATE $x'^{s}_{t_i} \leftarrow x^{s}_{t_i} +\tau(x'^{stylized}_{t_i}- x^{s}_1)$
   \STATE $v(x'^{stylized}_{t_i},t_i,\psi) \leftarrow -v_\theta(x^{c}_{t_i},t_i,\psi)+v_\theta(x'^{s}_{t_i},t_i,\psi)$
   \STATE $\hat{v}(x_{t_i}^{stylized},t_i,\psi) \leftarrow \frac{1}{2}\big(v(x^{stylized}_{t_i},t_i,\psi)+v(x'^{stylized}_{t_i},t_i,\psi)\big)$
   \STATE $x^{stylized}_{t_{i+1}} \leftarrow x^{stylized}_{t_i} + (t_{i+1}-t_{i})\hat{v}(x_{t_i}^{stylized},t_i,\psi)$
   \STATE $x'^{stylized}_{t_{i+1}} \leftarrow x'^{stylized}_{t_i} + (t_{i+1}-t_{i})\hat{v}(x_{t_i}^{stylized},t_i,\psi)$
   \ENDFOR
   \STATE {\bfseries Return:} $x^{stylized}_1 =  \frac{1}{2}(x_{t_{n_{max}}}^{stylized} + {x'}_{t_{n_{max}}}^{stylized})$
\end{algorithmic}
\end{algorithm}

\begin{algorithm}[tb]
   \caption{Dual-side Rectified Inversion-free Style-Transfer V2}
   \label{alg:dual-side}
\begin{algorithmic}
   \STATE {\bfseries Input:} 
   content image $x_1^c$, style image $x_1^s,\big\{t_i\big\}^{n_{\text{max}}}_{i=0}, n_{\text{max}},$ scaling factor $\tau=1$ and $\lambda=\frac{1}{2}$, 
   \STATE {\bfseries Output:} stylized image $x^{stylized}_1$
   \STATE {\bfseries Init:} $x^{stylized}_{t_0}=x^{c}_1,\quad x'^{stylized}_{t_0}=x^{s}_1$
   \FOR{$i=0$ {\bfseries to} $n_{\text{max}}$}
   \STATE $x_0^s \leftarrow x_0^c,\quad x_{0}^s \sim \mathcal{N}(0,\,1)$
   \STATE $x^{c}_{t_i} \leftarrow (1-t_i)x^c_1 + t_ix_0^c,\quad x^{s}_{t_i} \leftarrow (1-t_i)x^s_1 + t_ix_0^s$
   \STATE $x'^{c}_{t_i} \leftarrow x^{c}_{t_i} +\tau(x^{stylized}_{t_i}- x^{c}_1),\quad x'^{s}_{t_i} \leftarrow x^{s}_{t_i} +\tau(x'^{stylized}_{t_i}- x^{s}_1)$
   \STATE $x'_{t_i}=\lambda x'^s_{t_i} + (1-\lambda)x'^c_{t_i}$ // e.g., linear interpolation as the simplest combination, e.g., midpoint in this case
   \STATE $v(\hat{x}_{t_i}^{stylized},t_i,\psi) \leftarrow \Big(-v_\theta(x'_{t_i},t_i,\psi)+v_\theta(x'^{c}_{t_i},t_i,\psi)\Big)+\Big(-v_\theta(x'_{t_i},t_i,\psi)+v_\theta(x'^{s}_{t_i},t_i,\psi)\Big)$
   \STATE $x_{t_{i+1}}^{stylized} \leftarrow x_{t_i}^{stylized} + (t_{i+1}-t_{i})v(\hat{x}_{t_i}^{stylized},t_i,\psi)$
   \STATE ${x'}_{t_{i+1}}^{stylized} \leftarrow {x'}_{t_i}^{stylized} + (t_{i+1}-t_{i})v(\hat{x}_{t_i}^{stylized},t_i,\psi)$
   \ENDFOR
   \STATE {\bfseries Return:} $x^{stylized}_1 =  \frac{1}{2}(x_{t_{n_{max}}}^{stylized} + {x'}_{t_{n_{max}}}^{stylized})$
\end{algorithmic}
% \AddNote{top}{bottom}{right}{Optionally average $n_{\text{avg}}$ samples}
\end{algorithm}

\subsection{Attention Injection}
To guide the trajectory direction and enhance the incorporation of style cues (such as texture, color, and stroke) for improved stylized outcomes, we further introduce an attention injection into the transfer process. Given the noisy style image $x_t^s$, content image $x_t^c$ and stylized image $x_t^{stylized}$ (i.e., the midpoint of $x_t^c$ and $x_t^s$ in Fig. \ref{fig:framework_3}), the query($Q$), key($K$) and value($V$) using learnable projection matrices are as follows:
\begin{equation}
\begin{aligned}
&Q_t^s = W_qx_t^s; K_t^s = W_kx_t^s; V_t^s = W_vx_t^s,\\
&Q_t^{stylized} = W_qx_t^{stylized}; \\
&K_t^{stylized} = K_t^s; V_t^{stylized} = V_t^s.
\end{aligned}
\end{equation}
We replace the $K_t^{stylized}$ and $V_t^{stylized}$ with $K_t^{s}$ and $V_t^{s}$, enabling the model to query relevant style cues directly from the stylized image, thus enhancing style integration.
% by
% \begin{equation}
% \begin{aligned}
% x_{t+1}^{stylized} = Attn(Q_t^{stylized},K_t^{s},V_t^{s})
% \end{aligned}
% \end{equation}
% where $Attn(\cdot)$ is the attention layer the in the generator.
\section{Experiment}

\begin{figure*}
\centering
\includegraphics[width=1.0\linewidth]{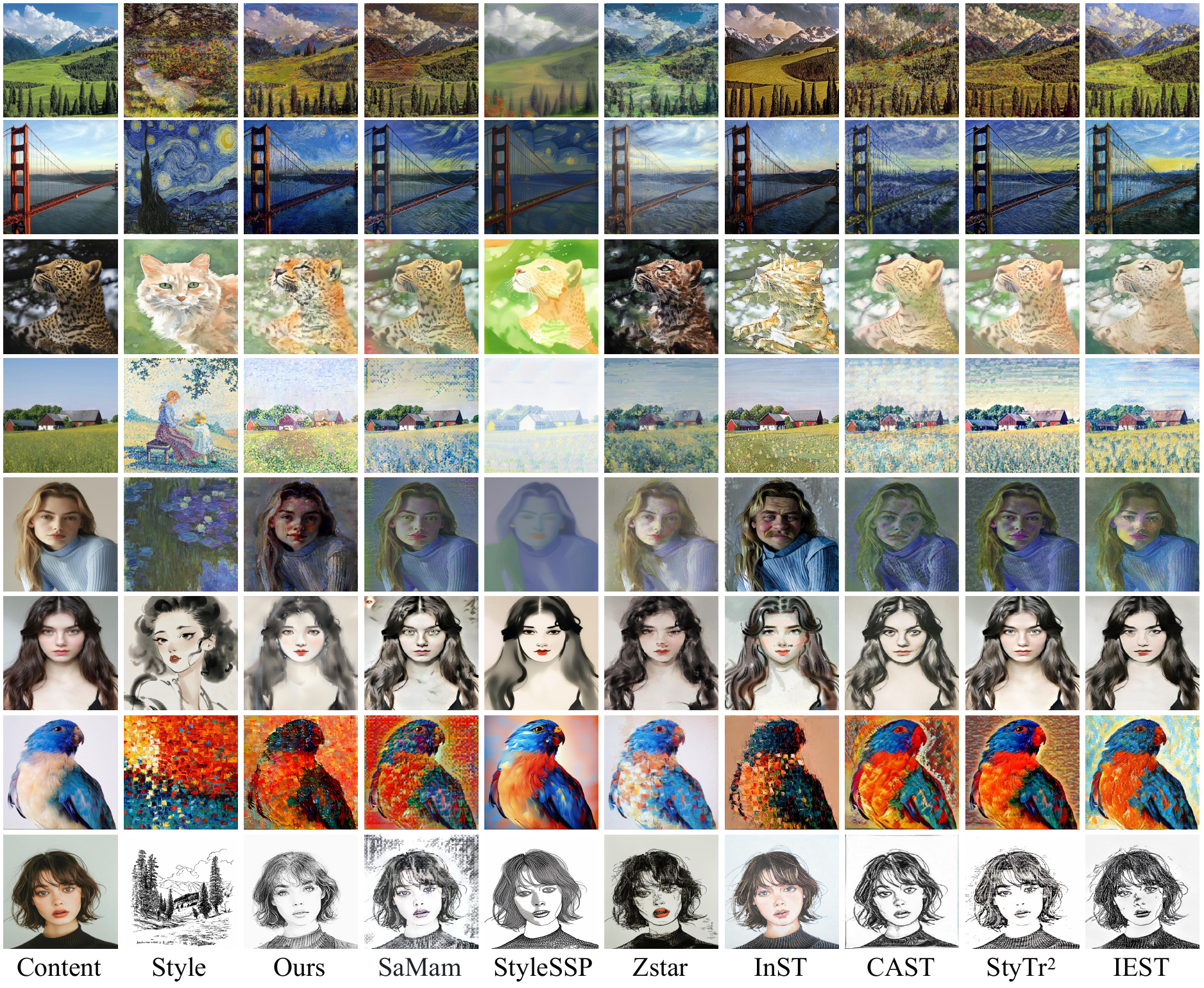}
\caption{Qualitative comparison of various style transfer methods. The first column displays the content images, the second column shows the style images, and the subsequent columns present the stylized results produced by different methods.
}
\label{fig:experiment}
\end{figure*}

\subsection{Implementation Details}
We conduct all experiments using the pre-trained Stable Diffusion 3 framework~\cite{EsserKBEMSLLSBP24}. The denoising process is set to $50$ steps, with a target guidance scale of $13.5$, and both content and style guidance scales set to $3.5$. All experiments are performed on an NVIDIA P40 GPU.

\subsection{Evaluation}
% We compare the proposed approach with several state-of-the-art methods, including diffusion-based methods (StyleSSP, Zstar~\cite{Deng_2024_CVPR}, and InST~\cite{zhang:2023:inversion})—and traditional style transfer methods （SaMam~\cite{2025cvprsamam}, CAST~\cite{zhang:2022:domain}, IEST~\cite{chen:2021:iest}, and STyTr$^2$~\cite{Deng:2022:CVPR}).
We perform comprehensive comparisons with state-of-the-art methods spanning different technical paradigms. The diffusion-based approaches include StyleSSP~\cite{StyleSSP}, Zstar~\cite{Deng_2024_CVPR}, and InST~\cite{zhang:2023:inversion}, which leverage pre-trained diffusion models for style transfer. Traditional style transfer methods encompass SaMam~\cite{2025cvprsamam}, CAST~\cite{zhang:2022:domain}, IEST~\cite{chen:2021:iest}, and StyTr$^2$~\cite{Deng:2022:CVPR}, representing recent advances in non-diffusion based approaches.

\begin{table*}[h]
\centering
\caption{Quantitative comparison of style transfer performance and inference speed. The most favorable outcomes are highlighted in \textbf{bold}, the second favorable outcomes are highlighted in \underline{underline}.}
\begin{tabular}{l|cccccccc}
\toprule
 Metric &Ours& SaMam &StyleSSP& Zstar&InST&CAST&StyTr$^2$ & IEST   \\
\midrule
$\contentloss \downarrow$ &  1.254&\underline{0.751}&1.274 &  1.591 & 1.323 & 1.292 &  \textbf{0.708} &  1.208 \\
\midrule
$\styleloss\downarrow$  & \textbf{3.033} &3.593&5.824&5.141&7.158 & 4.763& \underline{3.273}&3.436   \\
\midrule
$LPIPS \downarrow$  & \textbf{0.402} &0.421&0.563&0.477&0.615 & 0.564& \underline{0.416}&0.469   \\
\midrule
$FID \downarrow$  & \textbf{18.897} &21.633&21.092&20.751&\underline{19.755} & 22.362& 20.921&20.060   \\
\midrule
$ArtFID \downarrow$  & \textbf{27.792} &33.373&35.031&\underline{29.522}&30.664 & 36.426& 31.82&30.564   \\
\midrule
Infer. time \ (s/image) $\downarrow$ & 5.506 &0.031&50.012 &44.138&1445.000  &  \textbf{0.005} & 0.203&\underline{0.007}\\
\bottomrule
\end{tabular}
\label{tab:quancomps}
\end{table*}

\subsubsection{Qualitative Evaluation}

Figure~\ref{fig:experiment} presents a qualitative comparison with existing style transfer approaches.
StyTr$^2$ employs a pure transformer architecture to mitigate content leakage, yet it still exhibits insufficient style expression, as shown in the \nth{4} and \nth{5} rows.
IEST and CAST leverage contrastive learning to enhance style representation; however, the deep stylistic features are not fully captured, leading to style inconsistency and limited fidelity in reproducing artistic qualities, particularly in texture rendering (see the \nth{2} and \nth{4} rows).
InST learns a dedicated style embedding for each style image, but it occasionally captures deviant stylistic attributes (as in the \nth{4} and \nth{8} rows) and alters content structure (exemplified in the \nth{5} row).
Both Zstar and StyleSSP rely on accurate image inversion, which poses challenges for content preservation.
Specifically, Zstar leverages pre-trained diffusion representations in a training-free manner, yet it tends to either retain excessive content or distort content through style application (see the \nth{1}, \nth{5}, and \nth{8} rows).
StyleSSP introduces a sampling startpoint enhancement strategy to alleviate content leakage, but at times underrepresents the target style, as shown in the \nth{1}, \nth{3}, and \nth{5} rows.
SaMam adopts a Mamba-based architecture and explores a new paradigm for style transfer, yet it faces limitations similar to StyTr$^2$ in faithfully representing complex artistic styles.

In contrast, our method leverages a pre-trained ReFlow model to capture rich artistic representations and introduces a dual-side inversion-free strategy that preserves the original content structure.
% As a result, our approach achieves a better balance between content preservation and expressive style transfer.
% As illustrated in the \nth{3} and \nth{5} rows, the original content is rendered with appropriate stylistic strokes, such as ink washes and oil painting textures—while maintaining the distinctive characteristics of the reference artwork.
Therefore, our approach achieves superior performance in balancing content preservation and style expression. As demonstrated in the \nth{5} and \nth{7} rows, our method successfully renders original content with appropriate stylistic attributes, faithfully reproduces oil painting brushstrokes while preserving the distinctive artistic qualities of the reference artworks. Furthermore, our approach maintains consistent performance across diverse style types, from impressionistic brushwork to detailed textual patterns, demonstrating its robustness and generalization capability.

\subsubsection{Quantitative Evaluation}
This section presents a comprehensive quantitative comparison between the proposed method and existing state-of-the-art approaches to validate its effectiveness.

\textbf{Evaluation Matrix.} Inspired by ~\cite{A_Comprehensive_Evaluation}, we evaluate our method using evaluation matrices such as content loss, style loss, $LPIPS$ (Learned Perceptual Image Patch Similarity)~\cite{lpips}, $FID$ (Fréchet Inception Distance)~\cite{FID} and $ArtFID$~\cite{artfid}.

Content loss and style loss use a pre-train VGG19~\cite{VGG19} to extract perceptual features and compute the content and style differences between stylized image $\hat{ I_c}$ and content/style images $I_c/I_s$, as
\begin{equation}
\contentloss = \frac{1}{\layernumber}\sum_{ i=0 }^{\layernumber} \lVert \phi_i(\hat{I}_{c}) - \phi_i(\inputcontent)  \lVert_2,
\label{fun:contentloss}
\end{equation}
where $\phi_i(\cdot)$ denotes features extracted from the $i$-th layer in a pre-trained VGG19 and $\layernumber$ is the number of layers. The style perceptual loss $\styleloss$ is defined as
\begin{equation}
\begin{split}
\styleloss = \frac{1}{\layernumber}\sum_{ i=0 }^{\layernumber} &\lVert \mu( \phi_i(\hat{I}_{c})) - \mu (\phi_i(I_{s})) \lVert_2 \\
&+  \lVert v( \phi_i(\hat{I}_{c})) - v (\phi_i(I_{s})) \lVert_2,
\end{split}
\label{fun:styleloss}
\end{equation}
where $\mu(\cdot)$ and $v(\cdot)$ denote the mean and variance, respectively. Specifically, the content loss is computed using the features extracted from the $5$-th layer, while the style loss is calculated using the features from the $1$-st to $5$-th layers.

Since traditional style transfer methods are typically trained with content and style losses, we employ $LPIPS$ and $FID$ to ensure a fair comparison and avoid test bias. Specifically, $LPIPS$ quantifies content fidelity between the stylized and content images, while $FID$ assesses style fidelity by measuring the distributional similarity between the stylized outputs and the style image.
Additionally, we adopt $ArtFID$, a metric that strongly correlates with human judgment, for a more comprehensive evaluation of stylization quality, which is computed as
\begin{equation}
\begin{split}
ArtFID = (1 + LPIPS) \cdot (1 + FID).
\end{split}
\label{fun:styleloss}
\end{equation}

\begin{figure*}
\centering
\includegraphics[width=1.0\linewidth]{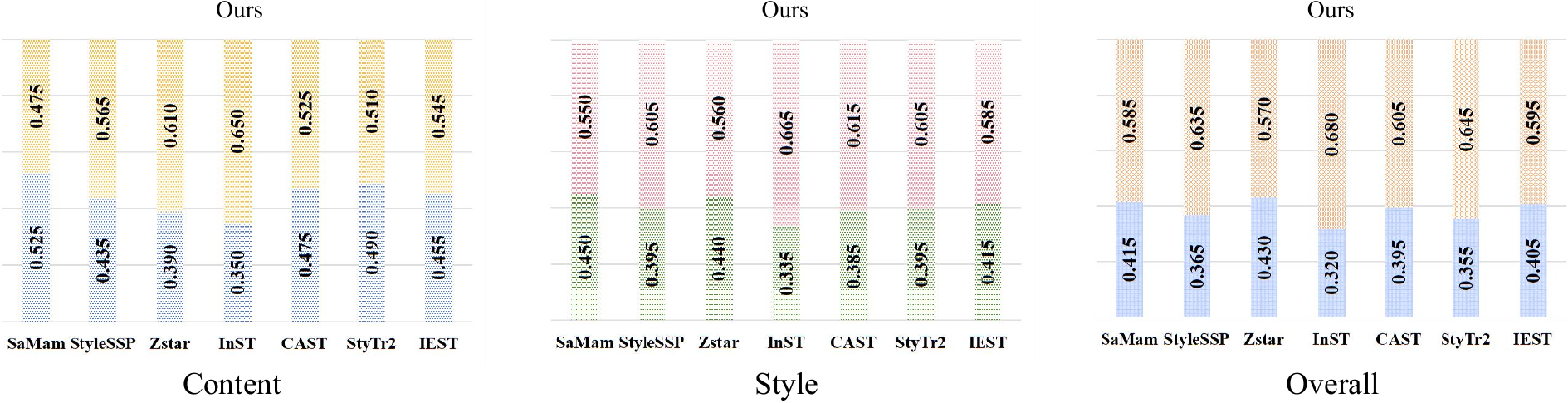}
\caption{User study results on content preservation, style fidelity, and overall quality. For each criterion, the upper segment of the bar represents the preference rate for our method, while the lower segment corresponds to the collective preference for comparison methods.
}
\label{fig:user_study}
\end{figure*}
\textbf{Quantitative Results.} To quantitatively compare different style transfer methods, we randomly selected $20$ style images from the WikiArt dataset and $20$ content images from the MS-COCO dataset, generating a total of $400$ stylized results for evaluation. The quantitative results are summarized in Table~\ref{tab:quancomps}.
As shown in the table, traditional style transfer methods such as SaMam and StyTr$^2$ achieve competitive performance in terms of $\contentloss$, indicating reasonable content preservation. However, they perform poorly on $FID$ and $ArtFID$, which reflects limitations in style quality and overall visual appeal. On the other hand, diffusion-based approaches like InST and Zstar attain higher $FID$ and $ArtFID$ scores, suggesting better style expression, yet they exhibit inferior content preservation as indicated by their $LPIPS$ and $\contentloss$ values.
In comparison, our method achieves the best performance in $LPIPS$, $FID$, $ArtFID$, and $\styleloss$, while ranking third in $\contentloss$. This demonstrates that our approach effectively balances content structure preservation and style fidelity, leading to visually consistent and high-quality stylization results.

\begin{figure*}
\centering
\includegraphics[width=1.0\linewidth]{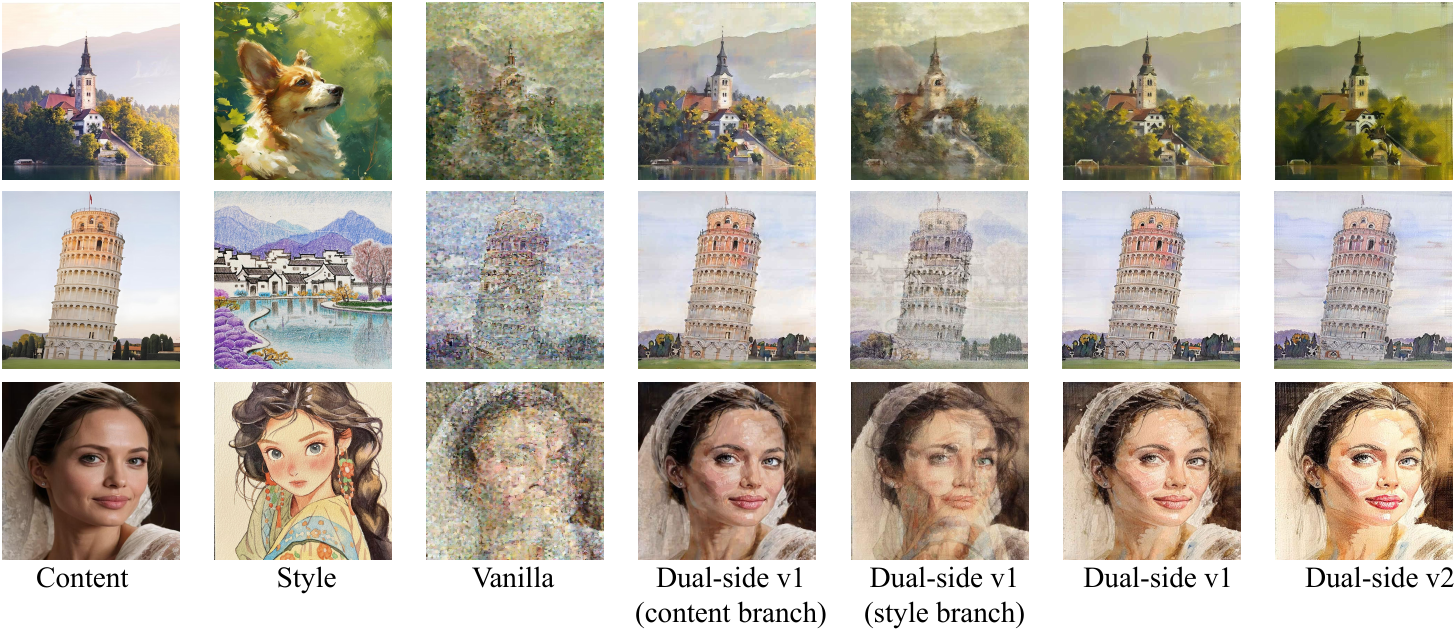}
\caption{Ablation study on inversion-free trajectories. The vanilla approach serves as a fundamental baseline, employing the velocity of noisy content and style images to generate the velocity for the stylized image. To distinctly demonstrate the impact of the content and style branches in our dual-path method, we present results using each branch independently, with the velocity of the other branch set to zero. The findings indicate that the content branch is instrumental in preserving content, while the style branch captures detailed brushwork, although elements of the style image are partially retained. When both branches are combined, the results more effectively illustrate the concept of balancing content and style within the same image. Lastly, the v2 version enhances the stylization outcome, as the velocity in the final steps is solely influenced by the stylized image. 
}
\label{fig:abl1}
\end{figure*}

\begin{figure}
\centering
\includegraphics[width=1.0\linewidth]{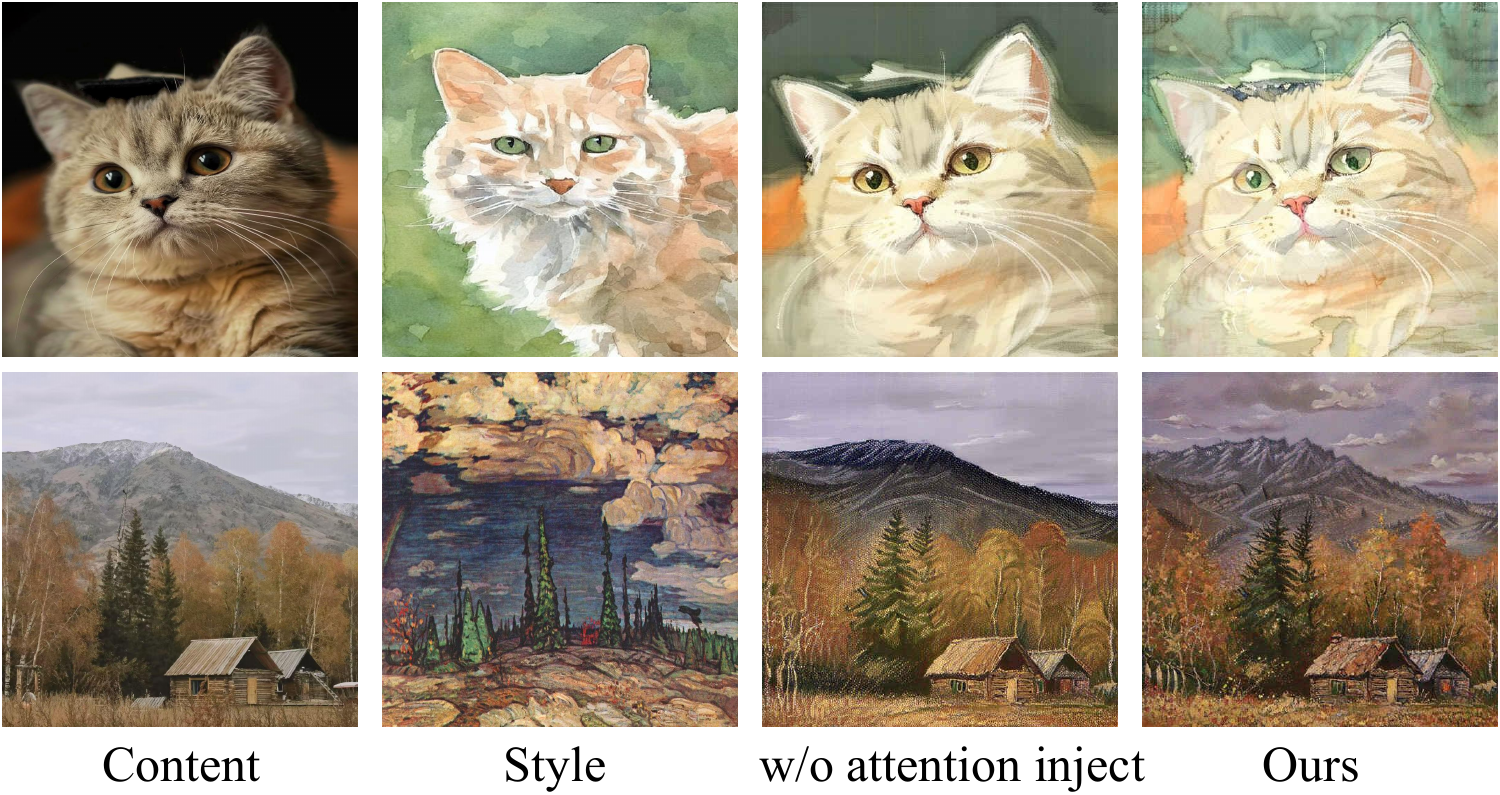}
\caption{Impact of the attention injection module. This module enhances style rendering while effectively preserving content information.
}
\label{fig:abl_attn}
\end{figure}

\textbf{Inference Speed.}
We compare the inference speed of different methods in TABLE~\ref{tab:quancomps}, with images uniformly set to a resolution of $512\times512$. Traditional style transfer methods, which employ a small network trained with predefined content and style losses, achieve the fastest inference owing to their single-step generation process. While our method is less efficient than these, it attains a considerably faster speed than diffusion-based approaches such as InST, StyleSSP, and Zstar. This efficiency gain is attributed to our inversion-free strategy and the straight denoising process of the ReFlow model. In conclusion, our approach maintains a favorable trade-off between stylization quality and computational efficiency.

\textbf{User Study.}
To obtain quantitative measurements of user preference and evaluate the subjective effectiveness of different stylization methods, we established a controlled experimental setup involving human participants. The study cohort consisted of $50$ participants with a demographic composition of $50\%$ male and $50\%$ female, spanning ages $18$-$45$ years, and representing both technical ($60\%$) and non-technical ($40\%$) backgrounds. The experimental images were selected from Quantitative Results incorporating results from our proposed method and various state-of-the-art comparison approaches. The experimental protocol required each participant to complete $28$ distinct evaluation groups, with each group presenting a randomly assigned content-style image pair together with the corresponding stylized results from our method and one randomly selected comparison method. To ensure consistent evaluation conditions, we implemented a fixed $30$-second viewing period for each image pair, displayed side-by-side with comprehensive zoom functionality to enable detailed examination. The presentation order of all evaluation questions was systematically randomized to control for potential sequence effects. During each evaluation group, participants provided comparative judgments across three specific dimensions: content preservation (evaluating which stylized result maintains the structural integrity and details of the original content image more effectively), style pattern quality (assessing which result demonstrates more convincing and aesthetically pleasing style characteristics), and overall visual effect (determining which result achieves superior overall aesthetic integration and visual appeal). 

This experiment resulted in the collection of 4,200 individual preference votes. We summarize the votes in Fig.~\ref{fig:user_study}, which indicates that our method delivers competitive performance across both content preservation and style representation metrics. Notably, its dominant performance in the overall effect evaluation confirms that it strikes an optimal trade-off, successfully integrating compelling style transfer with faithful content retention to yield superior visual quality.

\begin{figure*}
\centering
\includegraphics[width=1.0\linewidth]{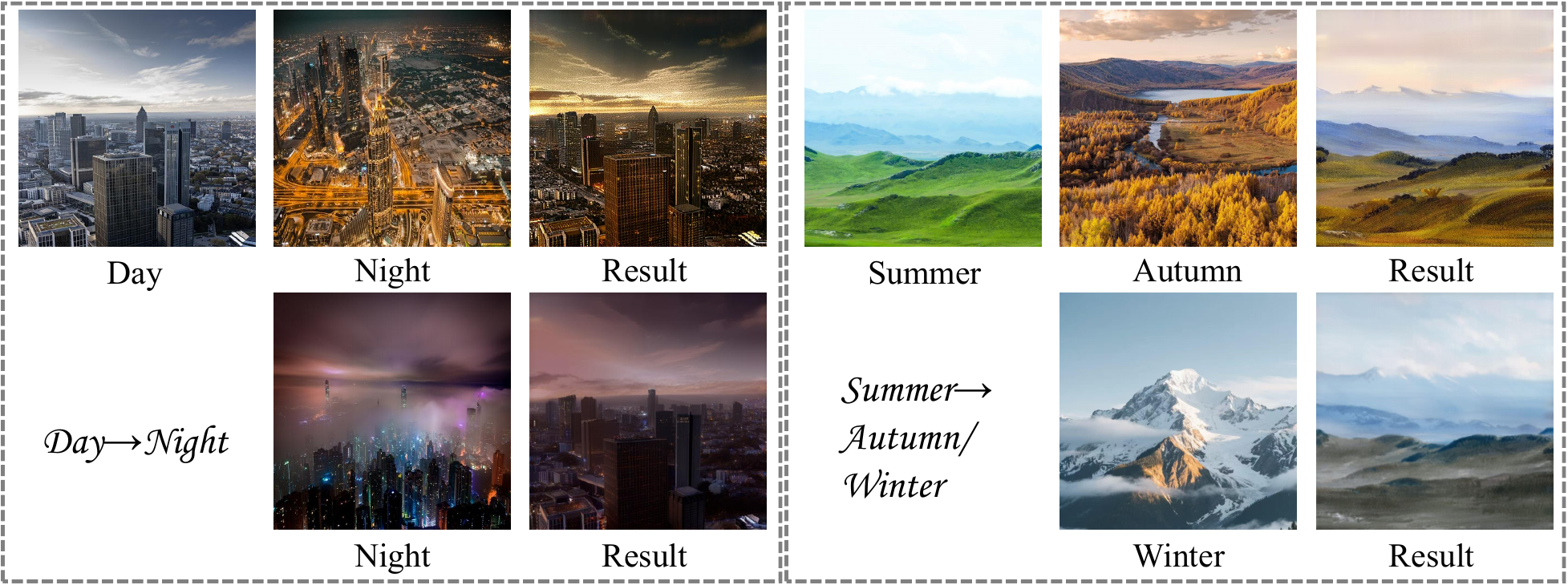}
\caption{Realistic style transfer results for time-of-day translation and seasonal alteration, demonstrating seamless shifts from day to night and summer to autumn/winter while preserving structural details and natural lighting and color transitions.}
\label{fig:real}
\end{figure*}

\subsection{Ablation Study}
\paragraph{{Influence of the inversion-free trajectories} }
In this section, we compare different inversion-free pipelines for style transfer: vanilla inversion-free style transfer, dual-side rectified inversion-free style transfer v1 only with content branch, dual-side rectified inversion-free style transfer v1 only with style branch, dual-side rectified inversion-free style transfer v1 with both side and dual-side rectified inversion-free style transfer v2. The comparison results are shown in Fig.~\ref{fig:abl1}.

Due to the direct combination of both content and style velocity, vanilla inversion-free style transfer produces noisy outputs, which also reflects the simple addition of both images. Content-branch rectified inversion-free style transfer strengthens content preservation, but overly suppresses style transfer, resulting in outputs that retain structure yet lack stylistic expressiveness.
Style-side rectified inversion-free style transfer, conversely, enhances style alignment by incorporating cues from the style reference, but yields overlap for content and style structure.
Dual-side rectified inversion-free style transfer v1 combines both strategies, improving overall coherence. However, due to the stylized images independent velocity term in Equation (\ref{eq:velocity_dual_side_v2_vanilla}), it still fails to capture detailed textures and strokes.
In comparison, our method our dual-side rectified inversion-free style transfer v2 overcomes those weaknesses. This enables superior preservation of content structure while faithfully reproducing rich, intricate stylistic elements, outperforming all prior variants.
\paragraph{Influence of the attention injection}
To demonstrate the effectiveness of our attention injection mechanism, we conduct an ablation study by removing it and comparing the resulting stylized images with those generated using the full framework. As illustrated in Fig.~\ref{fig:abl_attn}, the absence of attention injection leads to suboptimal style transfer outcomes. Specifically, while our dual-side inversion-free transfer process can still produce a basic stylized result, it fails to capture intricate style details, such as the subtle watercolor smudges, fine textures, and nuanced color gradients.
Therefore, attention injection enhances the model's ability to query and integrate precise style cues from the stylized image, improving overall visual coherence.

\subsection{Realistic Style Transfer}
Beyond artistic stylization, our method proves highly effective for realistic style transfer, a subfield concerned with applying real-world stylistic changes, such as time-of-day translation (day to night) or seasonal alteration (summer to autumn/winter). Realistic style transfer requires not only texture substitution but also a deep semantic understanding of how global attributes like lighting and color palette interact with scene content. As shown in Fig.~\ref{fig:real}, by using dual rectified flows, our approach successfully manipulates these holistic attributes, as it does not rely on a fragile inversion step and maintains a harmonious balance between content preservation and style application. This makes it particularly suitable for generating photorealistic and contextually appropriate stylization.

\begin{figure}
\centering
\includegraphics[width=1.0\linewidth]{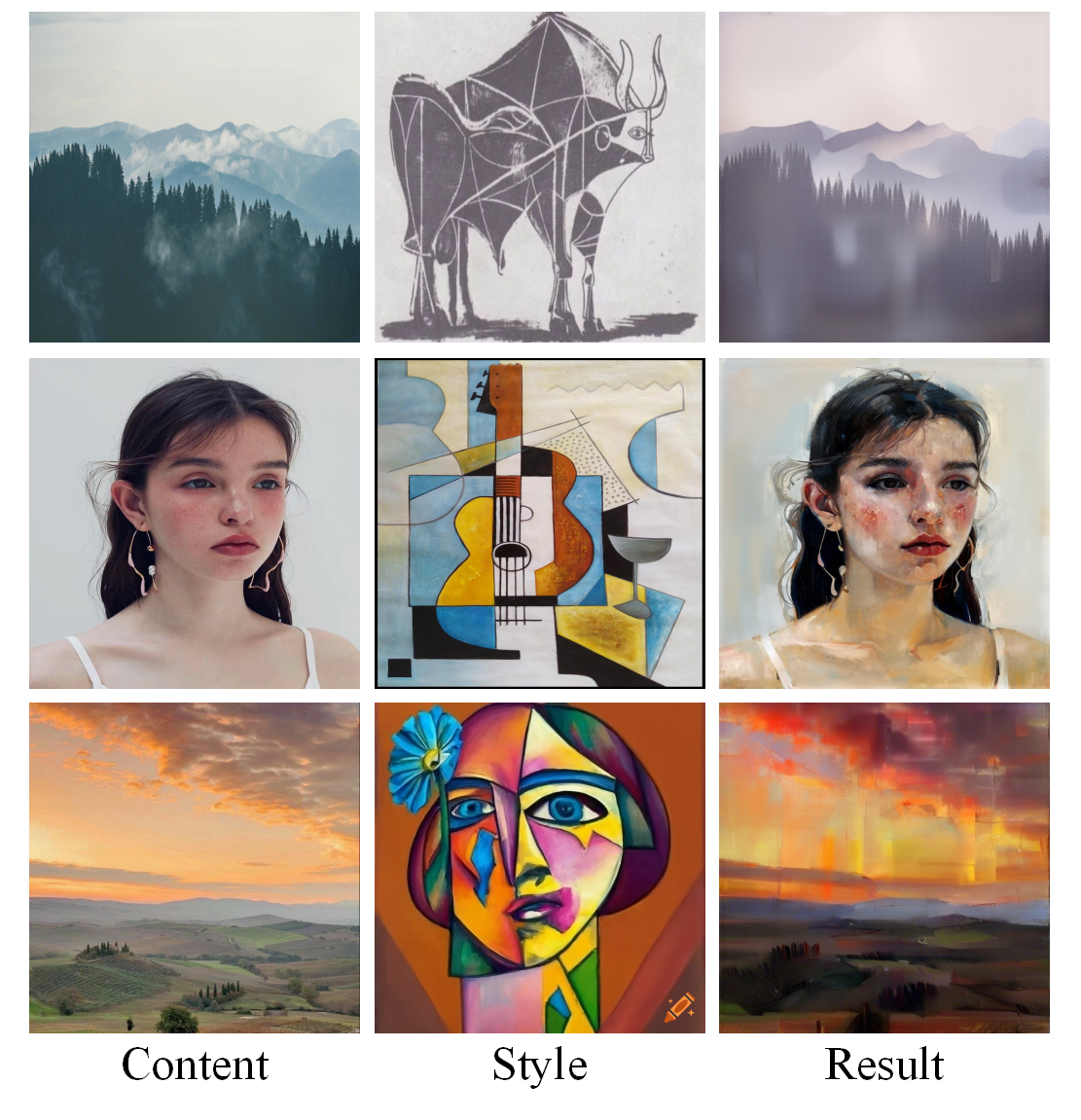}
\caption{Bad cases using cubist or abstract artworks, where the model captures the colors of cubist/abstract art but does not alter the shapes, resulting in insufficient stylization.
}
\label{fig:limitation}
\end{figure}

\subsection{Limitation}

While our method achieves compelling results in a wide range of arbitrary style transfer tasks, its limitations become apparent with abstract styles that require a semantic-level reinterpretation of content shapes, such as those found in Cubist or highly surrealistic artworks. As shown in Fig.~\ref{fig:limitation}, our model successfully imparts the style's color scheme and textural qualities but does not alter the underlying object geometries of the content image. This is an intentional trade-off rooted in our design choices that prioritize content structure preservation, which simultaneously prevents the content distortion common in other methods but also limits its expressivity for shape-deforming styles. The core challenge lies in the difficulty of disentangling and controlling geometric attributes within the existing generative model framework. Future investigations will focus on incorporating shape-aware constraints or exploring disentangled representations that can separately manipulate texture, color, and geometry, thereby extending the method's applicability to a broader spectrum of artistic expression.

\section{Conclusion}
In this work, we present a novel inversion-free style transfer framework that addresses the key limitations of training-free methods, which typically depend on unstable and computationally demanding inversion processes into noisy latent spaces, often yielding suboptimal stylization. By leveraging the advantages of ReFlow models and relying exclusively on forward ODE processes, our dual-flow approach effectively merges content preservation with style integration in parallel trajectories. This design eliminates the inherent computational overhead and instability of inversion-based techniques while introducing a midpoint interpolation heuristic to dynamically fuse velocities, ensuring robust and coherent outputs without resorting to simplistic content-style overlays. Additionally, our attention injection module further refines style cue incorporation, enhancing the overall artistic fidelity.
In conclusion, our method delivers high-fidelity stylized results with superior visual coherence, closely emulating the intuitive, holistic interplay of content and style in artistic creation.

% \section*{Acknowledgments}
% This should be a simple paragraph before the References to thank those individuals and institutions who have supported your work on this article.

 
\bibliographystyle{IEEEtran}
\bibliography{inv-free-style-transfer}

\vfill

\end{document}